\documentclass[10pt,twocolumn,letterpaper]{article}

\usepackage{iccv}
\usepackage{times}
\usepackage{epsfig}
\usepackage{graphicx}
\usepackage{amsmath}
\usepackage{amssymb}

\usepackage{color}
\usepackage{soul}
\usepackage{xcolor}
\usepackage{multirow}
\usepackage{booktabs}
\usepackage{enumerate}
\usepackage{enumitem}
\usepackage{pifont}%

\usepackage[pagebackref=true,breaklinks=true,letterpaper=true,colorlinks,bookmarks=false]{hyperref}

\usepackage[capitalize]{cleveref}
\crefname{section}{Sec.}{Secs.}
\Crefname{section}{Section}{Sections}
\Crefname{table}{Table}{Tables}
\crefname{table}{Tab.}{Tabs.}

\iccvfinalcopy 

\def\iccvPaperID{9} 

\ificcvfinal\fi

\begin{document}

\title{POSTER: A Pyramid Cross-Fusion Transformer Network for Facial Expression Recognition}

\author{Ce Zheng, Matias Mendieta,  Chen Chen\\
Center for Research in Computer Vision, University of Central Florida\\
{\tt\small \{ce.zheng,matias.mendieta\}@ucf.edu, chen.chen@crcv.ucf.edu}\\
}

\maketitle
\ificcvfinal\thispagestyle{empty}\fi

\begin{abstract}
  Facial expression recognition (FER) is an important task in computer vision, having practical applications in areas such as human-computer interaction, education, healthcare, and online monitoring. In this challenging FER task, there are three key issues especially prevalent: inter-class similarity, intra-class discrepancy, and scale sensitivity. While existing works typically address some of these issues, none have fully addressed all three challenges in a unified framework. In this paper, we propose a two-stream Pyramid crOss-fuSion TransformER network (POSTER), that aims to holistically solve all three issues. Specifically, we design a transformer-based cross-fusion method that enables effective collaboration of facial landmark features and image features to maximize proper attention to salient facial regions. Furthermore, POSTER employs a pyramid structure to promote scale invariance. Extensive experimental results demonstrate that our POSTER achieves new state-of-the-art results on RAF-DB (92.05\%), FERPlus (91.62\%), as well as AffectNet 7 class (67.31\%) and 8 class (63.34\%). \textcolor{magenta}{Code is available at \url{https://github.com/zczcwh/POSTER}}.
\end{abstract}

\section{Introduction}

Facial expression recognition (FER) is a critical task in computer vision, as it plays a crucial role in understanding human emotions and intentions. FER has various practical applications in fields such as human-computer interaction, education, healthcare, and online monitoring. Thus, it has received increasing interest in recent years, as highlighted in the comprehensive survey by Li et al. \cite{li2020survey}.


Traditional approaches \cite{zhao2007dynamic,zhong2012learning} utilize handcrafted features such as Histogram of oriented Gradients (HOG) \cite{HOG}, Local Binary Patterns (LBP) \cite{LBP}, and SIFT \cite{SIFT} for FER\cite{hog_fer,sift_fer,lbp_fer}. However, these handcrafted features are often not sufficiently robust and accurate.  More recent approaches have focused on leveraging the power of deep learning, as evidenced by the popularity of methods such as RAN \cite{wang2020RAN}, SCN \cite{wang2020SCN}, and KTN \cite{KTN}. These approaches have benefited from large-scale datasets that provide sufficient training data from challenging real-world scenarios and have shown significant performance improvement over traditional methods.

\begin{figure}[htp]
\vspace{-5pt}
  \centering
  \includegraphics[width=0.98\linewidth]{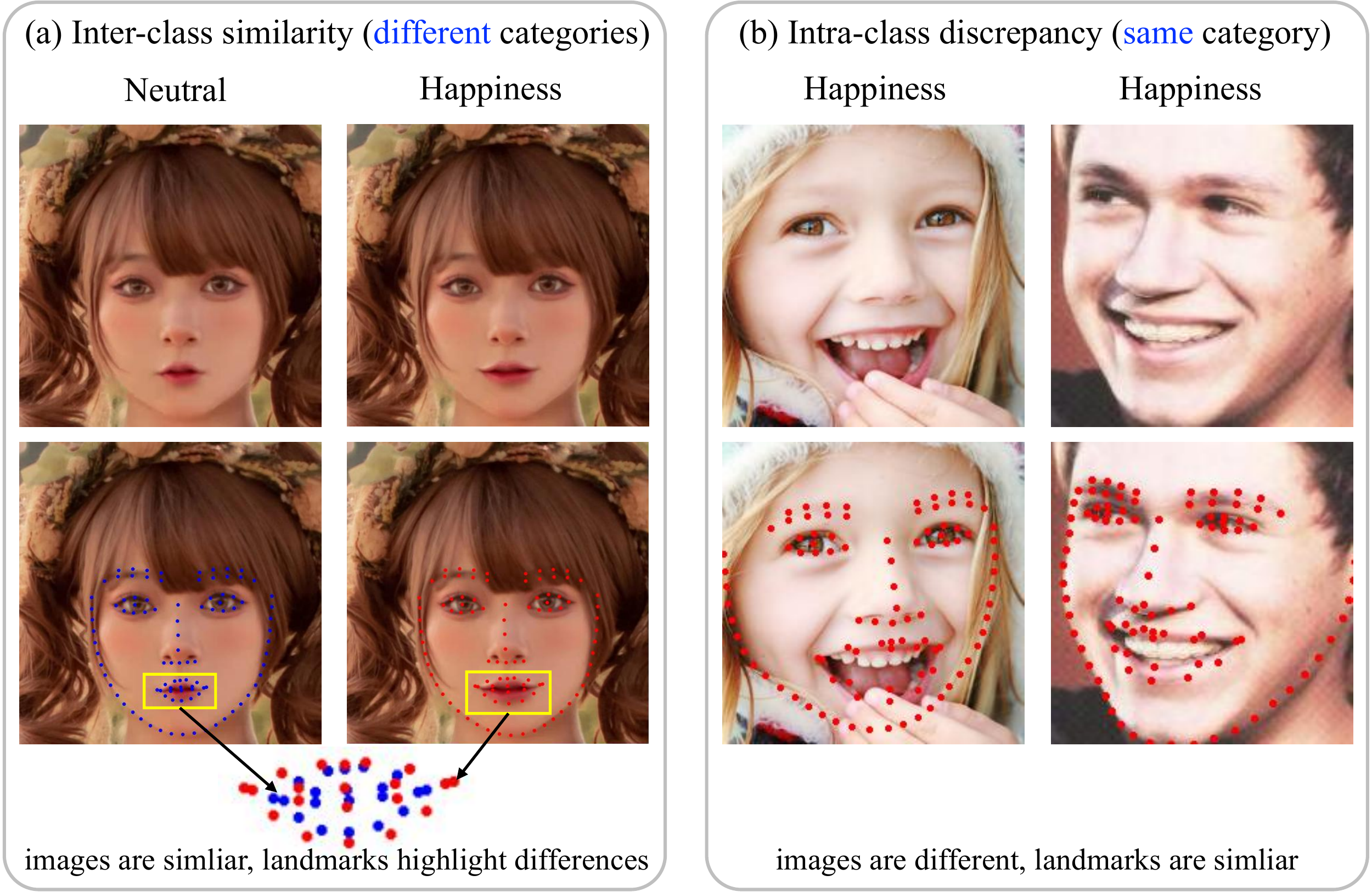}
  \caption{Inter-class similarity and intra-class discrepancy. (The facial landmarks are detected by \cite{PIPNet}.)}
    \label{fig:change}
  \vspace{-10pt}
\end{figure}

Despite the great progress made so far, there are still several challenges remain in FER: 
\setlist{nolistsep}
\begin{itemize}[noitemsep,leftmargin=*] 
  \item Inter-class similarity: Similar images with subtle changes between them can be classified into different expression categories. As illustrated in Fig.~\ref{fig:change} (a), A small change in a specific region of an image, such as the mouth, can determine the expression category, even when the overall appearance remains largely unchanged. Due to the subtlety of these differences, current methods may not be sufficiently robust to differentiate between such images.
  \item Intra-class discrepancy: Within the same expression category, images can have significant differences, such as the skin tone, gender, and age of a person varies across samples, as well as image background appearance. As shown in Fig.~\ref{fig:change} (b), two images both represent the expression of happiness but have very different visual appearances. 
  \item Scale sensitivity: When naively applied, deep-learning based networks are often sensitive to image quality and resolution. The image sizes within FER datasets and with in-the-wild testing images vary considerably. Therefore, ensuring consistent performance across scales is critical in FER \cite{vo2020psr}.
\end{itemize}

In light of these issues, some early works, some earlier works \cite{jung2015joint,hasani2017facial} (prior to 2018) attempted to improve the attention to detail and promote invariance to intra-class discrepancies by incorporating facial landmarks into their deep-learning based FER methods. These facial landmarks refer to a set of keypoints on the human face image that \textit{provide a sparse representation of key facial regions to complement direct image features}. However, these methods typically only use simple concatenation to combine the image and landmark features before the last fully connected layer or prior to a set of basic blocks. While incorporating facial landmarks with image features has potential to alleviate intra-class discrepancies and inter-class similarities, just simple concatenation is insufficient for exploring the correlations between landmarks and image features. Moreover, recent works \cite{FDRL,Xue_2021_transfer} address only intra-class discrepancy and inter-class similarity issues using image features, while another \cite{vo2020psr} addresses only scale sensitivity,  \textit{with none fully addressing all three challenges in FER}. Notably, landmark features, widely used in face-related tasks, have been largely ignored by recent FER approaches.

Therefore, in this paper, we propose a Pyramid crOss-fuSion TransformER network (POSTER) to tackle all three challenges of inter-class similarity, intra-class discrepancy, and scale sensitivity in FER. POSTER is a two-stream architecture that consists of an image stream and a landmark stream. Landmarks pinpoint salient regions and can serve as a guide for focusing feature attention, helping alleviate the inter-class similarity issue. As shown in Fig.~\ref{fig:change} (a), the salient region (mouth area) indicating the happiness expression is more easily captured through the discrepancy of landmarks rather than general image features. Also, the sparse landmark features can help to reduce the intra-class discrepancy because they are less sensitive in regard to various skin tones, genders, ages, and background appearance (image features may be affected significantly).
On the other hand, image features contain global information of the entire image that landmarks do not cover (such as cheeks, forehead, or a tears drop). Motivated by these intuitions, we propose to explore the correlations between landmark and image features with POSTER. Specifically, we design a transformer-based cross-fusion block that effectively allows the two streams to guide each other, and enables global correlation across features through attention. Experiments validate that the proposed cross-fusion transformer mechanism successfully alleviates inter-class similarity and intra-class discrepancy. 
Furthermore, to solve the scale-sensitivity for FER, we incorporate a pyramid architecture \cite{lin2017feature} with the cross-fusion transformer network to capture various resolutions of the extracted feature maps with different information granularities. \textbf{With our cross-fusion transformer design and integration of a feature pyramid structure, we fully address all three issues in a unified framework to bridge this research gap and achieve new SOTA results on several popular benchmarks.} 

Overall, our contributions are summarized as follows: 
\setlist{nolistsep}
\begin{itemize}[noitemsep,leftmargin=*] 
\item We propose a Pyramid crOss-fuSion TransformER network (POSTER) to alleviate inter-class similarity, intra-class discrepancy, and scale sensitivity issues in the FER. 

\item The cross-fusion transformer structure ensures that image features can be guided by landmark features with prior attention to salient facial regions, while landmark features can utilize global information provided by image features beyond landmarks.  

\item We extensively validate the efficiency and effectiveness of our proposed POSTER. We show that POSTER outperforms previous state-of-the-art methods on three commonly used datasets. (92.05\% on RAF-DB, 67.31\% on AffectNet, and 91.62\% on FERPlus.)
\end{itemize}

\section{Related Work}
\textbf{Deep learning in FER:} With the rapid progress of deep learning in computer vision, such techniques have been found increasingly useful for the challenging FER task \cite{zhang2021relative,zhang2022learn}. Wang et al. \cite{wang2020RAN} proposed a Region Attention Network (RAN) to capture facial regions for occlusion and pose variant FER. Farzaneh and Qi \cite{farzaneh2021DACL} introduced a Deep Attentive Center Loss (DACL) method to estimate the attention weight for the features for enhancing the discrimination. A sparse center loss was designed to achieve intra-class compactness and inter-class separation with the weighted features. Wang et al. \cite{wang2020SCN} proposed a Self-Cure Network (SCN) to suppress uncertainty, which prevents the network from overfitting incorrectly labeled samples. Shi et al. \cite{shi2021lARM} designed an Amending Representation Module (ARM) to reduce the weight of eroded features and decompose facial features to simplify representation learning.  

\textbf{Facial landmarks in FER:} Facial landmark detection aims to estimate the location of predefined keypoints on the human face. The detected facial landmarks are used in many face analysis tasks such as face recognition \cite{DeepFace,liu2017sphereface}, face tracking \cite{khan2017synergy}, and emotion recognition \cite{jung2015joint,hasani2017facial,qiu2019facial}. Significant progress has been made by employing deep learning techniques in the facial landmark detection task, and many accurate facial landmark detectors such as \cite{WangSCJDZLMTWLX19,wang2021face,chandran2020attention,PIPNet} have been proposed. Taking advantage of these off-the-shelf detectors, researchers have paid more attention to utilizing facial landmarks as informative features for the FER task. Jung et al. \cite{jung2015joint} proposed two networks, where one receives the image as input and another receives facial landmark as input. The output of the two networks is integrated by weighted summation. Hassani et al. \cite{hasani2017facial} proposed a 3D Inception-ResNet where facial landmarks are multiplied with image features at certain layers.
Khan \cite{khan2018facial} used the facial landmarks to crop small regions first, then generate features as the input for the neural networks. 
However, existing methods that utilize facial landmarks ignore the correlations of landmark features and image features. Among SOTA methods \cite{DMUE,shi2021lARM,Xue_2021_transfer} in the FER task, none of them use facial landmarks. 

\textbf{Vision transformer:} The breakthrough of transformer networks in Natural Language Processing (NLP) has sparked great interest in the computer vision domain. In NLP, the transformer is designed to model long sequence inputs. When adapting to the computer vision task, ViT \cite{Dosovitskiy2020ViT} split the image into patches, then the self-attention mechanism in the transformer can capture long-range dependencies across these patches. 
In FER, Aouayeb et al. \cite{aouayeb2021fer} directly applied the ViT structure by adding a SE block before the MLP-head for the FER task. Xue et al. \cite{Xue_2021_transfer} proposed a transformer-based method named TransFER. After extracting feature maps with a backbone CNN, the local CNN blocks were designed to locate diverse local patches. Then, a transformer encoder explored the global relationship between these local patches with a multi-head self-attention dropping module. 

Compared to previous works, our POSTER employs a two-stream pyramid cross-fusion transformer network to explore the correlation of image features and landmark features to tackle inter-class similarity, intra-class discrepancy, and scale sensitivity issues simultaneously in FER.




\section{Methodology}

\begin{figure*}[htp]
\vspace{-0.15cm}
  \centering
  \includegraphics[width=0.9\linewidth]{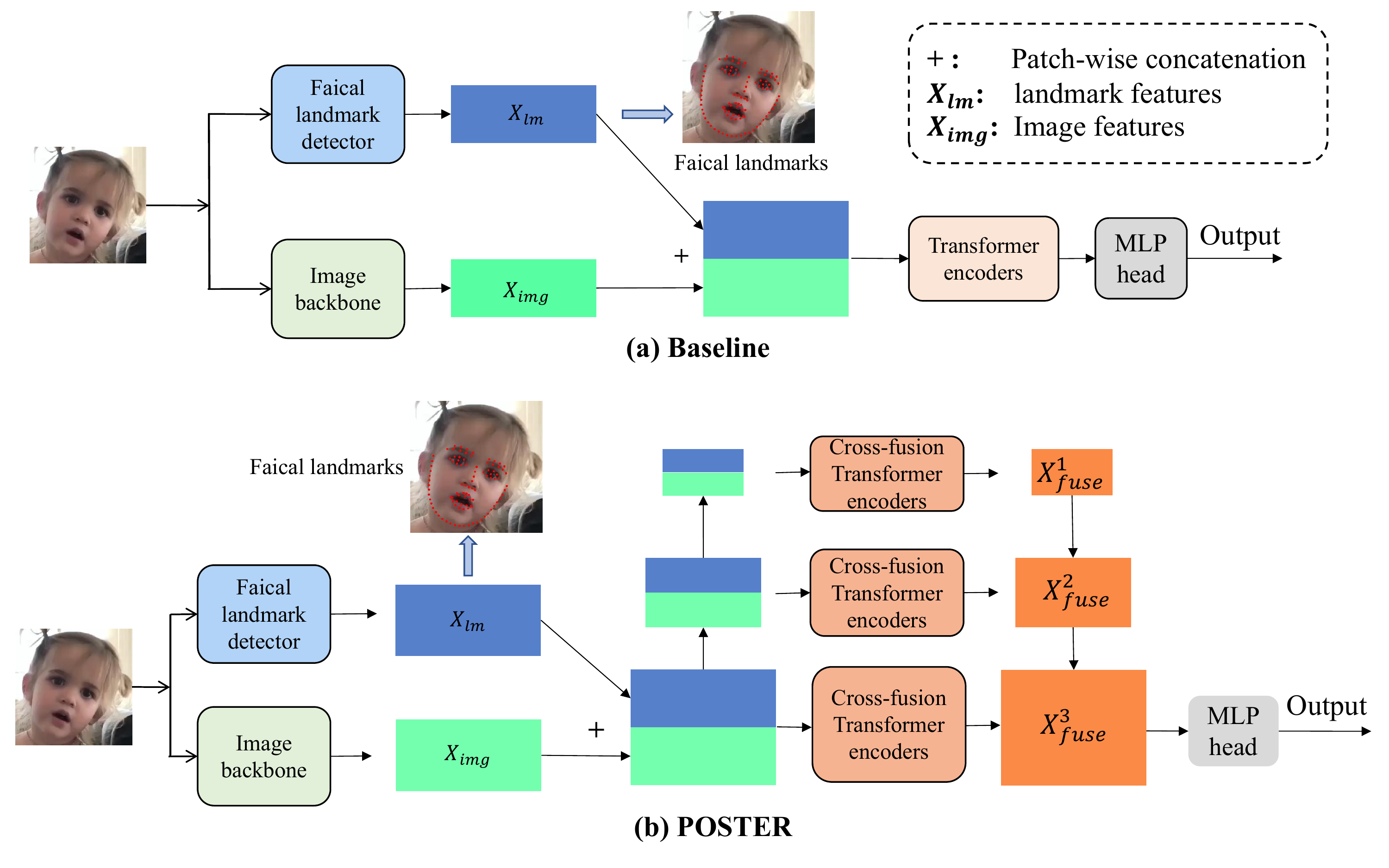}
  \vspace{-5pt}
  \caption{The architectures of baseline (a) and our proposed POSTER (b) for Facial Expression Recognition (FER). A facial landmark detector (MobileFaceNet \cite{PyTorchFaceLandmark}) is applied to obtain landmark features $X_{lm}$. The image backbone (IR50 \cite{ir50}) is used to extract image features $X_{img}$. ``\textbf{+}'' denotes patch-wise concatenation operation.}
  \vspace{-5pt}
    \label{fig:pipeline}
  \vspace{-5pt}
\end{figure*}

\begin{figure}[htp]
  \centering
  \includegraphics[width=0.99\linewidth]{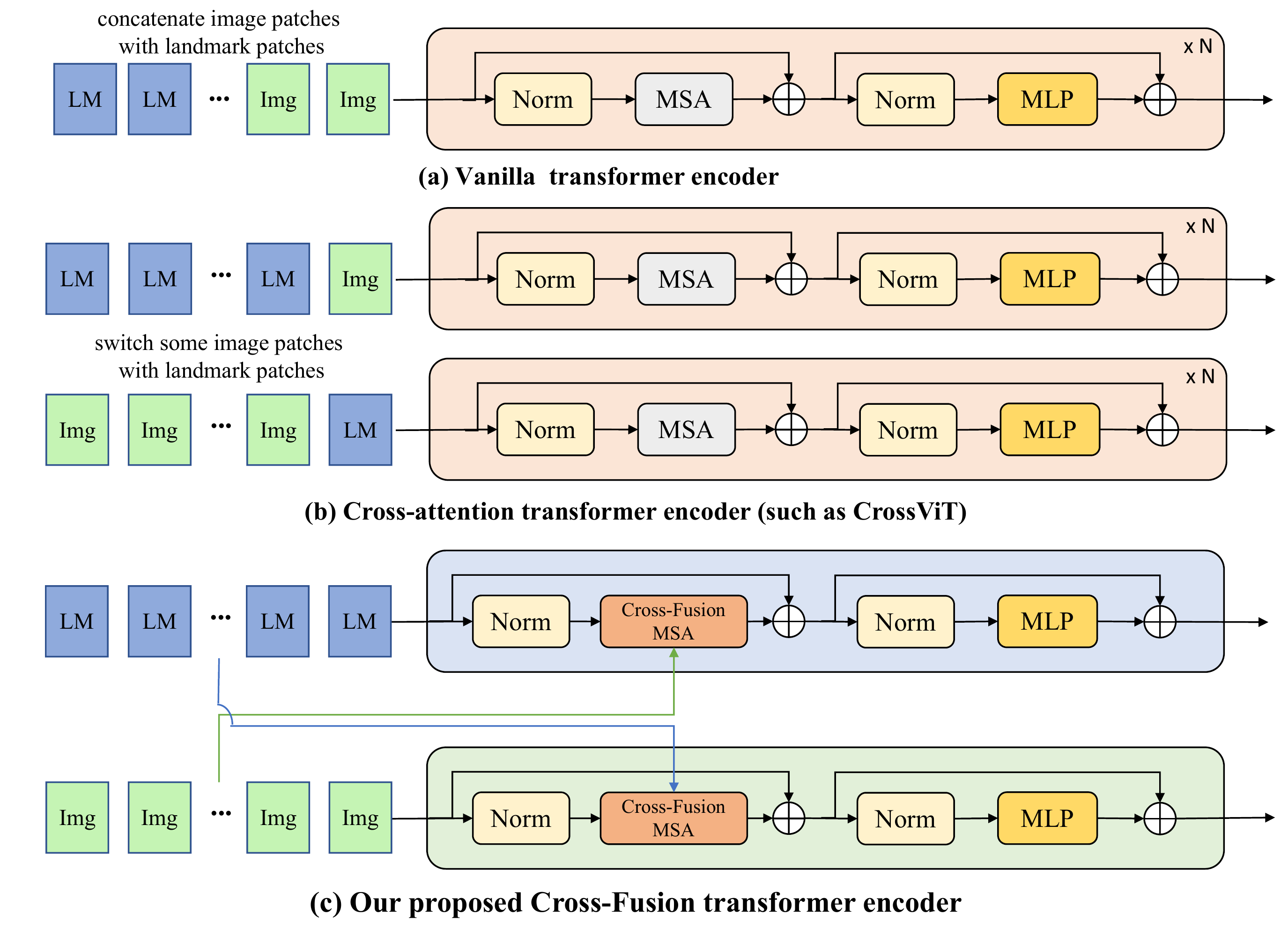}
  \caption{(a) The vanilla transformer encoder (from ViT \cite{Dosovitskiy2020ViT}). (b) The cross-attention transformer encoder (such as CrossViT) (c) Our proposed cross-fusion transformer encoder.}
  \vspace{-5pt}
    \label{fig:blocks}
  \vspace{-5pt}
\end{figure}

\subsection{Baseline} \label{Baseline}
First, we will describe our baseline architecture as shown in Fig.~\ref{fig:pipeline} (a).
We define an input image $X \in \mathbb{R} ^{H \times W \times 3}$, where $H$ and $W$ are the height and width of the image, respectively. In the network, we begin with an image backbone (e.g. IR50 \cite{ir50}) to get image features $X_{img} \in \mathbb{R} ^{P \times D}$. An off-the-shelf facial landmark detector (e.g. MobileFaceNe \cite{PyTorchFaceLandmark}) is used to obtain the landmark features  $X_{lm} \in \mathbb{R} ^{P \times D}$, where $P$ is the number of patches (same as the number of landmark keypoints) and $D$ is the feature dimension, respectively. During training, the image backbone is fine-tuned, while the off-the-shelf facial landmark detector is frozen to maintain proper landmark outputs.

After obtaining the image features $X_{img}$ and landmark features $X_{lm}$, an intuitive solution is to aggregate $X_{img}$ with $X_{lm}$ as the fused features $X_{fuse} \in \mathbb{R} ^{2P \times D}$ (concatenate in patch dimension  $P$). The transformer architecture utilizes its self-attention mechanism to capture the correlations across patches. Thus, we directly apply transformer encoders (illustrated in Fig.~\ref{fig:blocks} (a)) to operate on the fused features $X_{fuse}$. 

The self-attention mechanism is achieved by the Multi-head Self-Attention Layer (MSA) in the transformer architecture. The input $X_{fuse} \in \mathbb{R} ^{2P \times D}$ is first mapped to three matrices: the query matrix $Q$, key matrix $K$ and value matrix $V$ by three linear transformations: 
{\small
\begin{align}
    Q = {X_{fuse}}W_Q, \quad K = {X_{fuse}}W_K, \quad V = {X_{fuse}}W_V,
\end{align}}
\noindent where $W_Q$, $W_K$ and $W_V$ $\in \mathbb{R} ^{D \times D}$.

The vanilla transformer attention block is illustrated in Fig.~\ref{fig:cross} (a) can be described as the following mapping function: 
\begin{align}
\small
    {\rm Attention}(Q,K,V) = {\rm Softmax}(QK^\top/ \sqrt{d})V,
\end{align}
where $\frac{1}{\sqrt{d}}$ is the scaling factor for appropriate normalization to prevent extremely small gradients.

Next, the vanilla transformer encoder architecture consisting of MSA and MLP with a layer normalization operator is shown in  Fig.~\ref{fig:blocks} (a). The encoder output $X_{fuse \textunderscore out} \in \mathbb{R} ^{2P \times D}$ keeps the same size as the encoder input $X_{fuse} \in \mathbb{R} ^{2P \times D}$, and is represented as follows:
\begin{flalign}
\small
   & X'_{fuse} = {\rm MSA}(Q,K,V) + X_{fuse}, \\
   & X_{fuse \textunderscore out} = {\rm MLP}({\rm Norm}(X'_{fuse})) + X'_{fuse},
\end{flalign}

\noindent where  $\rm MSA(\cdot)$ represents the Multi-head Self-Attention block, $\rm Norm(\cdot)$ is the normalization operator, and $\rm MLP(\cdot)$ denotes the multilayer perceptron. 
Finally, an MLP head returns the predicted emotion label $Y \in \mathbb{R} ^{N}$ where $N$ is the number of classes.

Although the transformer encoders can inherently model the image features and landmark features jointly to some extent with simple concatenation of $X_{img}$ with $X_{lm}$, the strong correlations between image features and landmark features are not fully exploited by the vanilla transformer since features are concatenated together. In the next section, we describe POSTER which improves the correlation and representation capability of the model.

\subsection{POSTER} \label{POSTER}

\begin{figure}[htp]
\vspace{-0.15cm}
  \centering
  \includegraphics[width=0.99\linewidth]{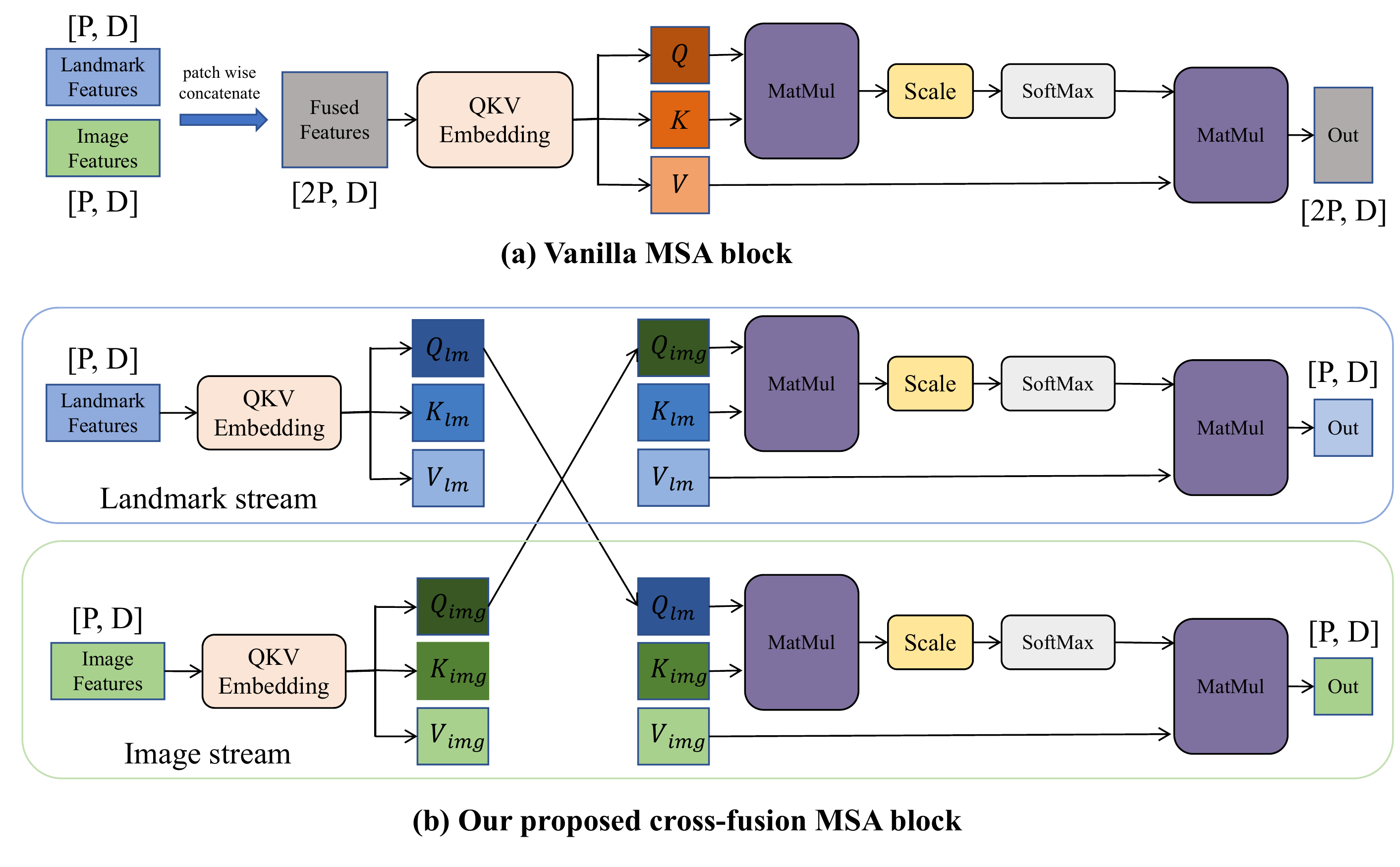}
  \vspace{-5pt}
  \caption{(a) The vanilla MSA block in transformer encoder. (b) Our proposed cross-fusion MSA block in cross-fusion transformer encoder.  P denotes the number of patches and D is the embedding dimension.}
    \label{fig:cross}
  \vspace{-5pt}
\end{figure}

The facial landmarks locate a set of keypoints in the human face region, which pinpoints salient regions related to facial expression. At the same time, global information beyond landmarks is also important for recognizing human expression (e.g. cheeks, wrinkled forehead). Motivated by this, we design a two-stream network that consists of both an image and a landmark stream. We swap the key matrices of the two streams, creating a cross-fusion operation to facilitate feature collaboration. Specifically, by performing this operation, we enable the image features to be guided by some prior knowledge of salient regions from the landmarks. Likewise, the representations of the landmark stream are provided with global context from the image features while moving through the block operations. In this way, we foster improved contextual understanding to alleviate intra-class discrepancy and inter-class similarity.
Therefore, given the extracted image features $X_{img}$ and landmark features $X_{lm}$, we design the cross-fusion MSA blocks as shown in Fig.~\ref{fig:cross} (b) to achieve our goal.

\noindent \textbf{Cross-fusion transformer encoder:}
For the MSA in the image stream, the input $X_{img} \in \mathbb{R} ^{P \times D}$ is mapped to three image matrices: query matrix $Q_{img}$, key matrix $K_{img}$ and value matrix $V_{img}$ by three linear transformations. For the landmark stream, the input $X_{lm} \in \mathbb{R} ^{P \times D}$ is mapped to three landmark matrices: query matrix $Q_{lm}$, key matrix $K_{lm}$ and value matrix $V_{lm}$ by three linear transformations: 
{\small
\begin{align}
    Q_{img} = {X_{img}}W_{Q1}, K_{img} = {X_{img}}W_{K1}, V_{img} = {X_{img}}W_{V1}, \\
    Q_{lm} = {X_{lm}}W_{Q2}, K_{lm} = {X_{lm}}W_{K2}, V_{lm} = {X_{lm}}W_{V2},
\end{align}
}
\noindent where $W_{Q1}$, $W_{Q2}$, $W_{K1}$, $W_{K1}$, $W_{V1}$ and $W_{V2}$ $\in \mathbb{R} ^{D \times D}$.

The cross-fusion transformer block illustrated in Fig.~\ref{fig:blocks} (b) can be described as the following mapping functions: 
\begin{align}
\small
    {\rm Attention_{(\it img)}} = {\rm Softmax}(Q_{lm}K_{img}^\top/ \sqrt{d})V_{img}, \\
    {\rm Attention_{(\it lm)}} = {\rm Softmax}(Q_{img}K_{lm}^\top/ \sqrt{d})V_{lm},
\end{align}
\noindent where $\frac{1}{\sqrt{d}}$ is the scaling factor for appropriate normalization to prevent extremely small gradients. The queries $Q_{img}$ and $Q_{lm}$ are swapped between the image and landmark streams. By doing this, the image features are empowered with salient regions provided by landmark features. On the other hand, the landmark features are provided with global information from the image features.

The cross-fusion transformer encoder is shown in  Fig.~\ref{fig:blocks} (b). The outputs of the cross-fusion transformer encoder $ X_{img \textunderscore out}$ and $ X_{lm \textunderscore out}$ given the image features $X_{img}$ and landmark features $X_{lm}$ are represented as follows:
\begin{flalign}
\small
   & X'_{img} = {\rm CFMSA_{img}}(Q_{lm},K_{img},V_{img}) + X_{img}, \\
   & X_{img \textunderscore out} = {\rm MLP}({\rm Norm}(X'_{img})) + X'_{img}, \\
   & X'_{lm} = {\rm CFMSA_{lm}}(Q_{img},K_{lm},V_{lm}) + X_{lm}, \\
   & X_{lm \textunderscore out} = {\rm MLP}({\rm Norm}(X'_{lm})) + X'_{lm}, 
\end{flalign}

\noindent where  $\rm CFMSA(\cdot)$ represents the Cross-Fusion MSA block in the image stream or the landmark stream, $\rm Norm(\cdot)$ is the normalization operator, and $\rm MLP(\cdot)$ denotes the multilayer perceptron. 

\noindent \textbf{Feature pyramid structure:} 
In FER, the image quality and resolution often vary considerably. Therefore, to support applicability across image scales, we adapt the well-known and effective technique of feature pyramid structure \cite{lin2017feature} for producing a multi-scale feature representation. 

Specifically, we construct large/medium/small levels of extracted features as shown in Fig.~\ref{fig:pipeline} (b). The large/medium/small features are fed to separate cross-fusion transformer encoders to capture specific feature scales. The output of the three cross-fusion transformer encoders is aggregated to form the emotion feature. Finally, an MLP head returns the predicted emotion label $Y \in \mathbb{R} ^{N}$, where $N$ is the number of classes.

\noindent \textbf{Distinction with other cross-attention design:} There are some methods proposed cross-attention transformer such as CrossViT \cite{chen2021crossvit}. The goal is to switch the patches from two different modalities as illustrated in Fig \ref{fig:blocks} (b). Different from this CrossViT-style transformer encoder, the motivation of our proposed cross-fusion transformer encoder is the feature collaboration. We enable the image features to be guided by some prior knowledge of salient regions from the landmarks.  Meanwhile, the representations of the landmark stream are provided with global context from the image features while moving through the block operations.

\noindent \textbf{Take-away:} Overall, we revitalize the idea of a two-stream architecture with our unique transformer-based cross-fusion method to alleviate inter-class similarity and intra-class discrepancy. Furthermore, we employ pyramid structures in POSTER to tackle scale-sensitivity, and therefore intelligently formulate a unified framework that addresses the three key challenges of FER. We verify the effectiveness of POSTER and its components in the following section.

\section{Experiments}
\subsection{Datasets}
\noindent \textbf{RAF-DB:} Real-world Affective Faces Database (RAF-DB) \cite{rafdb} is a large-scale facial expression dataset with 29,672 real-world facial images. All images are collected from the Internet with great variability in the subject's age, gender, ethnicity, lighting conditions, occlusions, etc. For the FER task, there are 15,339 facial expression images utilized (12,271 images are used for training and 3,068 images are used for testing) with seven basic expressions (happiness, surprise, sadness, anger, disgust, fear, and neutral). 

\noindent \textbf{FERPlus:} FERPlus \cite{ferplus} is extended from FER2013 \cite{fer2013} as used in the ICML 2013 Challenges. It is a large-scale dataset collected by APIs in the Google search engine. For the FER task, it contains 28,709 training images, 3,589 validation images, and 3,589 testing images. FERPlus relabeled the FER2013 into eight emotion categories (seven basic expressions plus contempt).  Following \cite{Xue_2021_transfer} and \cite{KTN}, we report the overall accuracy on the test set.

\noindent \textbf{AffectNet:} AffectNet \cite{affectnet} is one of the largest publicly available datasets for FER task. It is a large-scale in-the-wild dataset that contains more than 1M facial images collected from the Internet by querying three major search engines using 1250 emotion-related keywords in six different languages. Images are labeled into eight emotion categories (seven basic expressions plus contempt). 

\begin{table*}[]
\centering
  \caption{Comparison on RAF-DB, AffectNet, and FERPlus datasets. ``mean Acc'' denotes mean class accuracy and ``cls" is the abbreviation of classes. }
  \resizebox{0.8\linewidth}{!}{\begin{tabular}{cc|cc|cc|c}
\hline
\multicolumn{1}{l}{}          & \multicolumn{1}{l|}{} & \multicolumn{2}{c|}{RAF-DB} & \multicolumn{2}{c|}{AffectNet} & \multicolumn{1}{l}{FERPlus} \\ \hline
\multicolumn{1}{c|}{Method}  & Year                  & Acc         & mean Acc      & Acc (7 cls)  & Acc (8 cls) & Acc                         \\ \hline
\multicolumn{1}{c|}{SCN\cite{wang2020SCN}}      & CVPR 2020             & 87.03       & -             & -              & 60.23         & 89.39                       \\
\multicolumn{1}{c|}{PSR\cite{vo2020psr}}      & CVPR 2020             & 88.98       & 80.78         & 63.77          & 60.68         & -                           \\
\multicolumn{1}{c|}{RAN\cite{wang2020RAN}}      & TIP 2020              & 86.90       & -             & -              & -             & 89.16                       \\
\multicolumn{1}{c|}{DACL\cite{farzaneh2021DACL}}     & WACV 2021             & 87.78       & 80.44         & 65.20          & -             & -                           \\
\multicolumn{1}{c|}{KTN\cite{KTN}}      & TIP 2021              & 88.07       & -             & 63.97          & -             & 90.49                       \\
\multicolumn{1}{c|}{DMUE\cite{DMUE}}     & CVPR 2021             & 89.42       & -             & 63.11          & -             & -                           \\
\multicolumn{1}{c|}{FDRL\cite{FDRL}}     & CVPR 2021             & 89.47       & -             & -              & -             & -                           \\
\multicolumn{1}{c|}{ARM\cite{shi2021lARM}}      & arXiv 2021            & 90.42       & 82.77         & 65.20          & 61.33         & -                           \\
\multicolumn{1}{c|}{TransFER\cite{Xue_2021_transfer}} & ICCV 2021             & 90.91       & -             & 66.23          & -             & 90.83                       \\ 
\multicolumn{1}{c|}{Face2Exp\cite{zeng2022face2exp}} & CVPR 2022             & 88.54       & -             & 64.23          & -             & -                      \\
\multicolumn{1}{c|}{EAC\cite{zhang2022learn}} & ECCV 2022             & 89.99       & -             & 65.32          & -             & 89.64                     \\
\multicolumn{1}{c|}{FER-former\cite{li2023fer}} & arXiv 2023             & 91.30       & -             & -         & -             & 90.96                     \\
\hline
\multicolumn{1}{c|}{Baseline}     & -                     &  91.00      &  84.64         &  65.06          &  60.94        &  90.91   \\
\multicolumn{1}{c|}{POSTER}     & -                     &  \textbf{92.05}       &  \textbf{86.03}         &  \textbf{67.31}          &  \textbf{63.34}         &  \textbf{91.62}                       \\ \hline
\end{tabular}}
\label{tab: results}
\vspace{-5pt}
\end{table*}

\subsection{Implementation Details}
We implemented POSTER with Pytorch \cite{PyTorch} on two NVIDIA RTX 3090 GPUs.  in an end-to-end manner. We utilized IR50 \cite{ir50} as the image backbone, which is pretrained on Ms-Celeb-1M dataset \cite{Msceleb1m}. The weights of the image backbone are updated during training. For the facial landmark detector, we select MobileFaceNet \cite{PyTorchFaceLandmark} with all of the weights frozen to ensure it outputs landmark features. 
In the feature pyramid structure, we produce large/medium/small extracted features with embedding dim $D_H = 512$, $D_M = 256$, and $D_L = 128$, respectively. For the cross-fusion transformer encoders, each level of encoders consists of $depth = 8$ transformer encoders. The mlp ratio and drop path rate in transformer encoders are 2 and 0.01, receptively. We set the batch size to 100 with a learning rate of $4 \times 10^{-5} $. 
Unlike many methods \cite{farzaneh2021DACL,KTN} that rely on complicated loss, we use the standard label smoothing cross-entropy loss.

\subsection{Comparison with State-of-the-art Results}

\noindent \textbf{Evaluation on RAF-DB:}
Table~\ref{tab: results} compares POSTER with previous methods on the RAF-DB dataset. POSTER outperforms the SOTA methods both in terms of accuracy (the accuracy of all samples) and mean accuracy (the average of the accuracy of each category). Our POSTER yields the highest accuracy of 92.05 \%, which is 1.14 \% better than the second-best method (TransFER \cite{Xue_2021_transfer}). Our POSTER also achieves the highest score of 86.03 \% in mean accuracy, which is 3.26 \% better than the second-best method (ARM \cite{shi2021lARM}, as TransFER did not report the mean accuracy).   

\noindent \textbf{Evaluation on AffectNet:}
Table~\ref{tab: results} reports the results of POSTER with previous methods on the AffectNet dataset. AffectNet is the largest publicly available dataset with challenging facial expressions. The training set is extremely unbalanced (134,415 happiness images but only 3,803 disgust images; the gap is over 35$\times$). In terms of seven expression categories, our POSTER achieves the best accuracy of 67.31 \%, which is 1.08 \% higher than the second-best method (TransFER \cite{Xue_2021_transfer}). When considering the Contempt category (total eight classes), Our POSTER also yields the highest accuracy of 63.34 \%, which is 2.01 \% better than the second-best method (ARM \cite{shi2021lARM}).   

\noindent \textbf{Evaluation on FERPlus:}
Table~\ref{tab: results} evaluates POSTER with previous methods on FERPlus dataset. All images in FERPlus are grayscale images with small resolutions (48×48). Our POSTER still achieves the best accuracy of 91.62 \%, which is 0.79 \% higher than the second-best method (TransFER \cite{Xue_2021_transfer}). \ul{Overall, the superior performance on all three datasets has demonstrated the effectiveness of POSTER.}

\subsection{Results analysis}



\begin{table*}[]
\centering
  \caption{Class-wise accuracy on RAF-DB, AffectNet, and FERPlus datasets. The \textcolor{blue}{blue} color indicates the intra-class discrepancy has been reduced.}
  \resizebox{0.9\linewidth}{!}
  {
  \begin{tabular}{c|c|cccccccc|c}
\hline
                                                                             &          & \multicolumn{8}{c|}{cls Acc}                                            & \multirow{2}{*}{mean Acc} \\
Dataset                                                                      & Method   & Neutral & Happy & Sad   & Surprise & Fear  & Disgust & Anger & Contempt &                           \\ \hline
\multirow{2}{*}{RAF-DB}                                                      & Baseline & 90.44   & 96.71 & 90.38 & 87.23    & 62.16 & 76.25   & 88.89 & -        & 84.58                     \\
                                                                             & POSTER     &  \color{blue}92.35   &  \color{blue}96.96 &  \color{blue}91.21 &  \color{blue}90.27    &  \color{blue}67.57 &  75.00   &  88.89 & -        &  \color{blue}86.04                     \\ \hline
\multirow{2}{*}{\begin{tabular}[c]{@{}c@{}}AffectNet\\ (7 cls)\end{tabular}} & Baseline & 64.00       & 87.80     & 62.60     & 63.60        & 65.40     & 56.00       & 56.40     & -        & 65.11                         \\
                                                                             & POSTER     & \color{blue}67.20   & \color{blue}89.00 & \color{blue}67.00 & \color{blue}64.00    & 64.80 & 56.00   & \color{blue}62.60 & -        & \color{blue}67.23                     \\ \hline
\multirow{2}{*}{\begin{tabular}[c]{@{}c@{}}AffectNet\\ (8 cls)\end{tabular}} & Baseline & 57.20       & 74.60     & 61.20     & 63.00        & 63.40     & 61.40       & 52.00     & 54.71        & 60.94                         \\
                                                                             & POSTER     & \color{blue}59.40   & \color{blue}80.20 & \color{blue}66.60 & \color{blue}63.60    & \color{blue}63.60 & 59.80   & \color{blue}58.80 & 54.71    & \color{blue}63.34                     \\ \hline
\multirow{2}{*}{FERPlus}                                                     & Baseline & 92.52       & 95.52     & 91.88     & 85.60        & 91.08     & 50.00       & 62.79     & 13.33        & 72.84                         \\
                                                                             & POSTER     & \color{blue}93.26   & \color{blue}96.08 & \color{blue}92.39 & \color{blue}85.86    & \color{blue}91.45 & 43.75   & \color{blue}68.60 & \color{blue}33.33    & \color{blue}75.59                     \\ \hline
\end{tabular}
}
\label{tab: mean_acc_multi}
\end{table*}

\begin{table*}[]
\centering
  \caption{Prediction details given the target class on RAF-DB dataset. The \textcolor{blue}{blue} color indicates the intra-class discrepancy has been reduced. The \textcolor{red}{red} color denotes the inter-class similarity has been allivated. }
  \begin{tabular}{c|c|ccccccc}
\hline
         & {Ground truth} & \multicolumn{7}{c}{Prediction Percentage}                           \\ \cline{3-9} 
         &                         & Neutral & Happy & Sad   & Surprise & Fear  & Disgust & Anger \\ \hline
Baseline & Neutral                 & 90.44   & 2.50  & 4.56  & 1.62     & 0.00     & 0.74    & 0.15  \\
POSTER     & Neutral                 & \color{blue}92.35   &  \color{red}1.91   &  \color{red}4.26  &  \color{red}1.32     & 0.00     &  \color{red}0.15    &  \color{red}0.00     \\ \hline
Baseline & Fear                    & 2.70    & 5.41  & 10.81 & 14.86    & 62.16 & 2.70    & 1.35  \\
POSTER     & Fear                    & 4.05    &  \color{red}2.70  &  \color{red}9.46 &  \color{red}12.16    & \color{blue}67.57 & 2.70    & 1.35  \\ \hline
Baseline & Surprise                 & 5.47    & 0.91  & 0.91  & 87.23    & 1.52  & 1.82    & 2.13  \\
POSTER     & Surprise                 &  \color{red}2.74   &  0.91  &  \color{red}0.30  & \color{blue}90.27    & 2.43  & 1.82    &  \color{red}1.52  \\ \hline
\end{tabular}
\label{tab: cls_acc}
\vspace{-10pt}
\end{table*}

\begin{figure}[htp]
  \centering
  \includegraphics[width=1\linewidth]{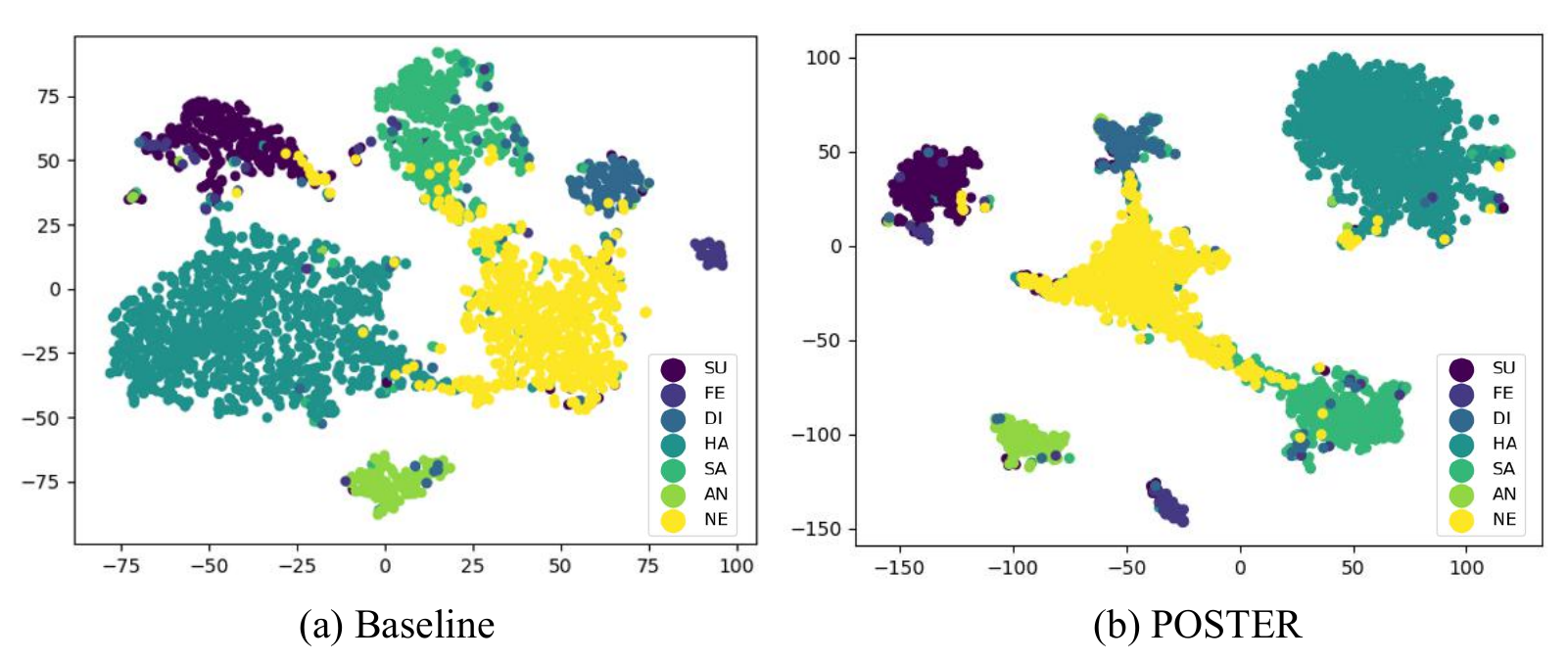}
  \caption{Visualisation of high dimensional features using t-SNE\cite{t_SNE} for the baseline and POSTER on the RAF-DB dataset.}
    \label{fig:tsne}
  \vspace{-5pt}
\end{figure}

In order to better understand the effectiveness of POSTER and its design, we perform further analysis on individual class performance.
Specifically, we first analyze the class-specific evaluations of our baseline and POSTER on RAF-DB, FERPlus, and AffectNet (7 classes and 8 classes) in Tables~\ref{tab: mean_acc_multi} and \ref{tab: cls_acc}.
For \textbf{RAF-DB}, the accuracy of Neutral, Happy, Sad, and Surprise categories of POSTER is higher than 90\% as shown in Table~\ref{tab: mean_acc_multi}. However, the accuracy of Fear category is relatively low. This is due to insufficient training samples on this category (281 of Fear images, while others such as Neutral, Happy, Sad, and Surprise are more than 1,000). Similar to the RAF-DB dataset, the accuracy of Disgust and Contempt categories in the \textbf{FERPlus} are relatively low as shown in Table~\ref{tab: mean_acc_multi}. This is also due to insufficient training samples on these two categories (only 116 of Disgust and 120 of Contempt, more than 20$\times$ less than other categories). For \textbf{AffectNet}, all images are collected from the internet, most of the training and testing samples are from in-the-wild settings. The extremely unbalanced training set and challenging in-the-wild samples make the performance quite lower than in RAF-DB and FERPlus datasets. The accuracy of Happy category achieves 80\% in Table~\ref{tab: mean_acc_multi} since there are 134,415 happiness samples (almost 50 \% of the total training images).

When comparing the class-wise accuracy in Table~\ref{tab: mean_acc_multi}, POSTER significantly increases the accuracy of most of the categories on three datasets, which indicates that POSTER reduces intra-class discrepancy for FER (marked in \textcolor{blue}{blue} color if the class accuracy has been improved). In Table~\ref{tab: cls_acc}, we show the percentage of prediction given the target categories. For example, given a total of 680 Neutral testing images, 628 images are correctly classified as Neutral by POSTER (92.35 \%) and 13 images are classified incorrectly as Happy (1.91 \%).  The percentage of the wrong prediction into certain categories also decreases as marked in \textcolor{red}{red} color, which shows that POSTER alleviates intra-class discrepancy for FER. Moreover, we visualize the high dimensional features before final output of our baseline and POSTER by t-SNE \cite{t_SNE}. The improved separation across classes and density with class clusters also verifies that POSTER alleviates the intra-class discrepancy and inter-class similarity.

\subsection{Ablation Study}
We conduct the ablation study on RAF-DB and AffectNet datasets to verify the contribution of the proposed structures and the impact of hyperparameters on performance. More ablation studies are provided in the \textcolor{blue}{supplementary}.  

\begin{figure}[htp]
  \centering
  \includegraphics[width=0.99\linewidth]{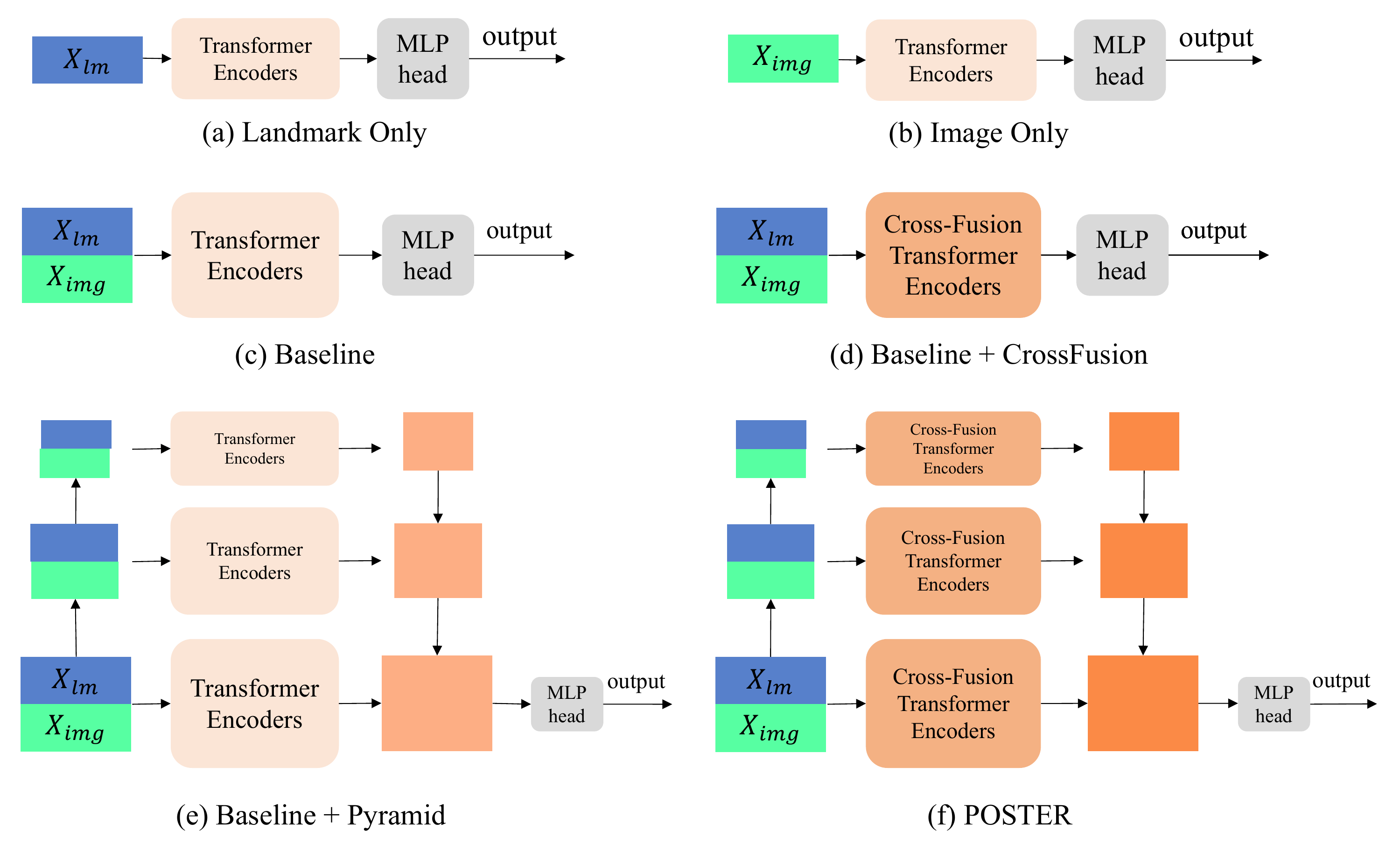}
  \caption{(a) Only use landmark features $X_{lm}$ for classification. (b) Only use image features $X_{img}$ for classification. (c) Our baseline in Sec.~\ref{Baseline}. (d) Our baseline with replacing vanilla transformer encoders with proposed cross-fusion transformer encoders. (e) Our baseline with adding a pyramid structure. (f) Our proposed POSTER in Sec.~\ref{POSTER}.  }
    \label{fig:ab_pipeline}
\end{figure}

\begin{table}[htp]
\centering
  \caption{ Ablation study on the proposed components.}
  \resizebox{0.99\linewidth}{!}
  {\begin{tabular}{c|cc|cc}
\hline
                           & \multicolumn{2}{c|}{RAF-DB}     & \multicolumn{2}{c}{AffectNet}   \\ \hline
Components                 & Acc            & Acc(mean)      & 7 cls          & 8 cls          \\ \hline
(a) Landmark only               & 80.08          & 72.21          & 49.88          & 45.34          \\
(b) Image only              & 90.51          & 82.73          & 64.95          & 56.60          \\
(c) Baseline               & 91.00          & 84.64          & 65.06          & 60.94          \\
(d) Baseline+pyramid       & 91.27          & 85.66          & 66.50          & 62.36          \\
(e) Baseline+cross\_fusion & 91.63          & 85.01          & 65.35          & 61.87          \\
(f) POSTER                   & \textbf{92.05} & \textbf{86.03} & \textbf{67.31} & \textbf{63.34} \\ \hline
\end{tabular}}
\label{tab: Ab_components}
\end{table}


\begin{figure*}[htp]
  \centering
  \includegraphics[width=0.99\linewidth]{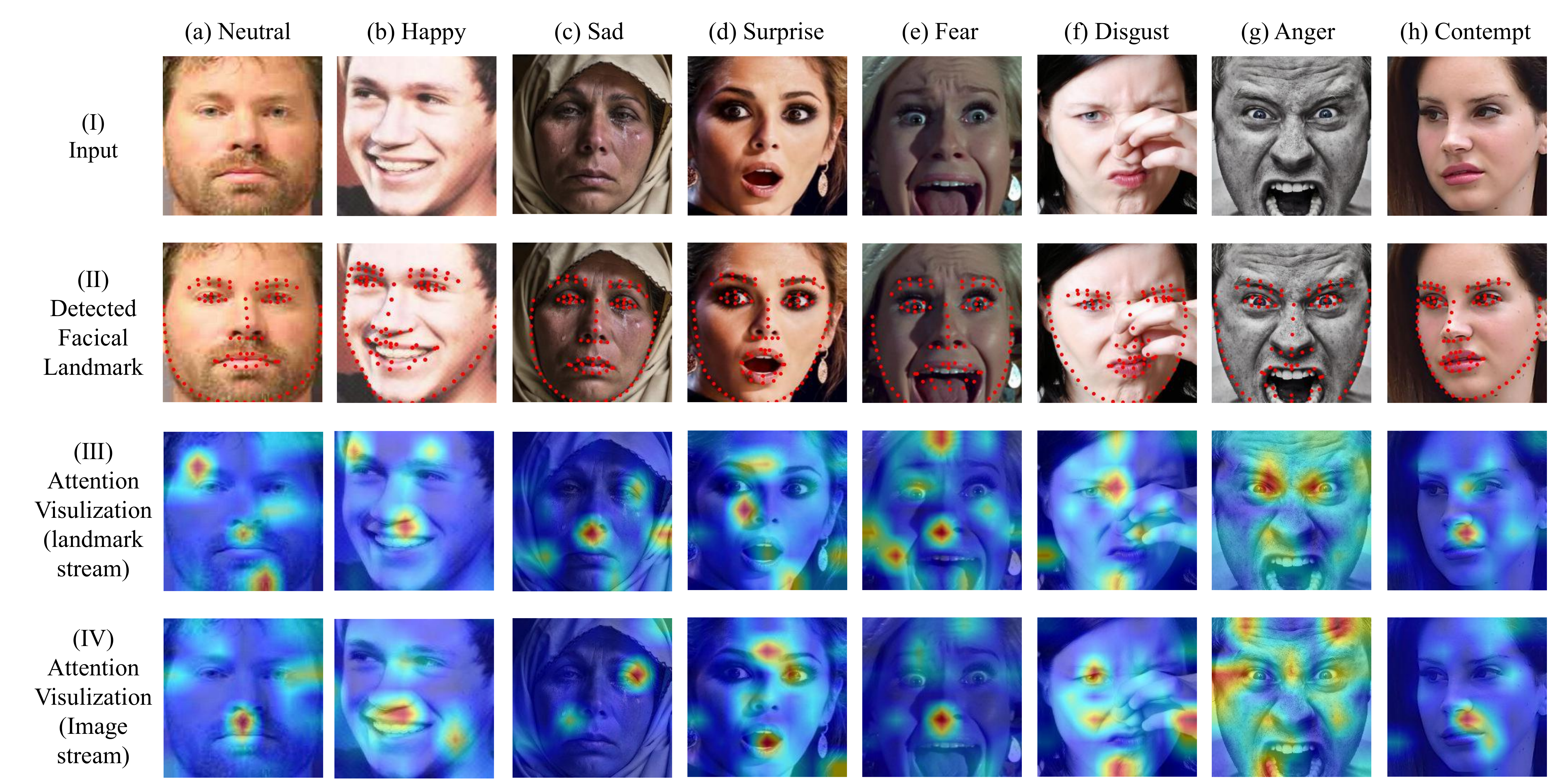}
  \caption{Attention visualization on facial images of different categories (images are from the AffectNet dataset). }
    \label{fig:attn_vis}
  \vspace{-5pt}
\end{figure*}

\noindent \textbf{Effectiveness of the architecture design:}
We investigate the different architectures of utilizing image features and landmark features and report the results in Table~\ref{tab: Ab_components}. In Fig.~\ref{fig:ab_pipeline} (a) and (b), we only use the landmark features and image features, respectively. Utilizing image features achieves better results than landmark features. When concatenating image features with landmark features as a baseline in  Fig.~\ref{fig:ab_pipeline} (c) (discussed in Sec. \ref{Baseline}), the performance is better than image features only and landmark features only. When replacing the vanilla transformer attention blocks with our proposed cross-fusion transformer attention blocks as shown in Fig.~\ref{fig:ab_pipeline} (d), the performance can be boosted, which verifies the effectiveness of our proposed cross-fusion design. Next, we report the performance with our pyramid structure illustrated in  Fig.~\ref{fig:ab_pipeline} (e). When comparing (c) with (d), and (e) with (f), the performance has improved because the pyramid structure can alleviate scale sensitivity issues given various input sizes and quality of training samples. Finally, our POSTER in Fig.~\ref{fig:ab_pipeline} (f) achieves the best results.


\noindent \textbf{Attention Visualization:} In order to visualize the parts of the facial image that contributes to the category clarification, we apply the visualization method of \cite{trans_visualization} to visualize the attention maps in the transformer. The relevance score is computed for each attention head in each layer of the transformer encoders, and these scores are then integrated by incorporating both relevancy and gradient information. More details are discussed in \cite{trans_visualization}.

The attention visualizations and detected facial landmarks for a set of facial images are shown in Fig.~\ref{fig:attn_vis}. The images of 8 categories are selected from the in-the-wild AffectNet dataset.
Since our proposed cross-fusion transformer attention block has two streams (landmark and image), we visualize the attention map of each stream separately. The attention learned by these two streams are different but complementary. 

For the landmark stream attention map, the highlight areas are commonly from the landmark areas. However, it also obtains some global attention beyond landmarks from the image features, such as cheek and forehead areas in (\romannumeral3, a),  (\romannumeral3, e), and  (\romannumeral3, h). For the image stream attention maps, the highlight areas indicate the discriminate patterns that led to a certain classification. For example, the tears drop in (\romannumeral4, c) is discovered, which indicates a sad expression. The forehead wrinkles in (\romannumeral4, g) are highlighted which relates to an angry expression. Because landmark features are cross-fused as guidance, the attention can focus on important regions related to landmarks. Moreover, some areas such as between the eyebrows and above the brow are also captured by image-guided features to improve the indication of the expression. By integrating attention from both landmark and image streams, POSTER can effectively recognize facial expressions with excellent performance.

\section{Conclusions}
In this paper, we have proposed a Pyramid crOss-fuSion TransformER network (POSTER) for the FER task. As a two-stream network, landmark features are detected by an off-the-shelf facial landmark detector, and image features are extracted by a backbone CNN. The correlations of image features and landmark features are fully exploited in POSTER by our proposed cross-fusion transformer architecture. POSTER tackles all three challenges of inter-class similarity, intra-class discrepancy, and scale sensitivity in FER. Extensive experiments on three commonly used FER datasets have demonstrated that POSTER outperforms state-of-the-art methods.  

\newpage

\appendix
\noindent \textbf{\large{Supplementary Material}}

\section{Overview}

The supplementary material contains more ablation studies, which are organized into the following sections:

\begin{itemize}
\item Section \ref{pyramid}: Pyramid Layers

\item Section \ref{cross}: Cross-fusion Mechanism

\item Section \ref{flops}: Model Size and FLOPs Comparison

\item Section \ref{depth}: Transformer encoders depth. 

\item Section \ref{confusion}: Confusion Matrices
\end{itemize}


\section{Pyramid Layers:}
\label{pyramid}
To investigate the optimal pyramid layers with embedded dimensions, we conduct the experiments on the RAF-DB dataset and the results are shown in Table \ref{tab: Ab_pyramid}. The three pyramid layers with embedded dimensions [512, 256, 128] achieve similar results to the four pyramid layers with embedded dimensions [512, 256, 128]. Considering the computational budget, POSTER adopts the three pyramid layers with embedded dimensions [512, 256, 128] as the optimal choice. 

\begin{table}[htp]
\tiny
\centering
  \caption{ Ablation study on the Pyramid Layers.}
  \resizebox{0.9\linewidth}{!}{
  \begin{tabular}{l|cc}
\hline
\multicolumn{1}{c|}{}   & \multicolumn{2}{c}{RAF-DB}         \\ \hline
layers                  & \multicolumn{1}{c|}{Acc}   & mAcc  \\ \hline
{[}512{]}               & \multicolumn{1}{c|}{91.63} & 85.01 \\ \hline
{[}512, 256{]}          & \multicolumn{1}{c|}{91.77} & 85.49 \\ \hline
{[}512, 256, 128{]}     & \multicolumn{1}{c|}{92.05} & 86.03 \\ \hline
{[}512, 256, 128, 64{]} & \multicolumn{1}{c|}{92.04} & 85.97 \\ \hline
\end{tabular}}
\vspace{-5pt}
\label{tab: Ab_pyramid}
\end{table}

\begin{table}[htp]
\centering
  \caption{ Ablation study on Cross-fusion Mechanism.}
  \resizebox{0.9\linewidth}{!}{
  \begin{tabular}{l|cc}
\hline
\multicolumn{1}{c|}{}          & \multicolumn{2}{c}{RAF-DB}                             \\ \hline
                               & \multicolumn{1}{c|}{Acc}   & mAcc                      \\ \hline
No swap                        & \multicolumn{1}{c|}{91.27} & 85.66                     \\ \hline
swap for the first block       & \multicolumn{1}{c|}{91.68} & 85.71                     \\ \hline
swap for the first two blocks  & \multicolumn{1}{c|}{91.89} & 85.88                     \\ \hline
swap for the first four blocks & \multicolumn{1}{l|}{91.91} & \multicolumn{1}{l}{85.86} \\ \hline
swap for all 8 blocks          & \multicolumn{1}{c|}{92.05} & 86.03                     \\ \hline
\end{tabular}}
\label{tab: Ab_crossfusion}
\vspace{-5pt}
\end{table}

\section{Cross-fusion Mechanism}
\label{cross}
For POSTER, we swap $Q_{img}$ and $Q_{lm}$ for cross fusion during transformer attention for each MSA block. Image features can be guided by some prior knowledge of salient regions from the landmarks. Likewise, the landmark features are provided with additional global context from the image features. In this way, we foster improved contextual understanding to alleviate intra-class discrepancy and inter-class similarity. 
We also conduct experiments to evaluate different  cross-fusion mechanisms in Table \ref{tab: Ab_crossfusion}. Based on the results, swapping $Q_{img}$ and $Q_{lm}$ for all blocks achieves the best performance. 

\begin{table*}[htp]
\small
\centering
  \caption{Comparison on Parameters and FLOPs. The image backbone (IR50) and facial landmark detector (MobileFaceNet) are taken into account when computing Params and FLOPs of POSTER-T, POSTER-S, and POSTER. }
  \resizebox{0.8\linewidth}{!}
  {\begin{tabular}{cccccc}
\hline
Methods     & Year      & Params & FLOPs  & Acc(RAF-DB)   & Acc(AffectNet)\\ \hline
DMUE\cite{DMUE}        & CVPR 2021 & 78.4M  & 13.4G  & 89.42         & 63.11\\
TransFER\cite{Xue_2021_transfer}    & ICCV 2021 & 65.2M  & 15.3G  & 90.91         & 66.23  \\\hline
POSTER-T &     -     & 52.2M  & 13.6G  & 91.36         & 66.86 \\
POSTER-S &     -     & 62.0M  & 14.7G  & 91.54         & 67.13 \\
POSTER   &     -     & 71.8M  & 15.7G  & 92.05         & 67.31 \\ \hline
\end{tabular}}
\label{tab: flops}
\vspace{-5pt}
\end{table*}

\begin{table*}[htp]
\centering
\caption{The trade-off between the complexity versus the performance of POSTER.}
  \resizebox{\linewidth}{!}{
  \begin{tabular}{c|cc|cc|cc|cc}
\hline
           & \multicolumn{2}{c|}{}       & \multicolumn{2}{c|}{transformer blocks} & \multicolumn{2}{c|}{overall (with backbones)} & RAF-DB & AffectNet\_7cls \\
           & \# of blocks     & emb\_dim & Params(M)          & FLOPs (G)          & Params(M)             & FLOPs (G)            & Acc    & Acc       \\ \hline
Image only & 8 (single ViT)   & 512      & 19.7               & 2.4                & 50.8                  & 13.1                 & 90.51  & 65.35     \\ \hline
POSTER-T  & 4 (Cross-Fusion) & 512      & 19.7               & 2.4                & 52.2                  & 13.6                 & 91.36  & 66.86     \\ \hline
POSTER-S  & 6 (Cross-Fusion) & 512      & 29.5               & 3.5                & 62.0                  & 14.7                 & 91.54  & 67.13     \\ \hline
POSTER     & 8 (Cross-Fusion) & 512      & 39.3               & 4.5                & 71.8                  & 15.7                 & 92.05  & 67.31     \\ \hline
\end{tabular}
\label{tab: trade-off}
}
\end{table*}

\section{Model Size and FLOPs Comparison:}
\label{flops}
Previous methods did not pay much attention to the model's computational and memory complexity. The total number of parameters (Params) and floating-point operations (FLOPs) of the model are two key characteristics for a fair comparison, but are often neglected. Furthermore, many recent papers did not release their implementation code. Here we list the Params and FLOPs of DMUE \cite{DMUE} (estimated based on their released code) and TransFER \cite{Xue_2021_transfer} (provided by the author) compared with our POSTER in Table~\ref{tab: flops}. We introduce three versions of our POSTER: POSTER-T (tiny version, the depth of transformer encoders is 4),  POSTER-S (small version, the depth of transformer encoders is 6), and POSTER (the depth of transformer encoders is 8). The \textit{Params and FLOPs of the image backbone and landmark detector are included for our methods.} 

POSTER-T has lower Params and similar FLOPs compared with DMUE \cite{DMUE}, but POSTER-T has much better performance both on RAF-DB and AffectNet datasets. When comparing with TransFER \cite{Xue_2021_transfer},  POSTER-T outperforms TransFER for all aspects. If the goal is to pursue higher performance, POSTER would be a good choice since computational and memory complexity is similar to other methods while achieving higher accuracy.

To investigate the trade-off between complexity versus performance, we show the complexity metrics in Table \ref{tab: trade-off}. 
When only using image features modeled by conventional transformer encoders, the Accuracy is 90.51 \% on RAF-DB and 65.35 \% on AffectNet\_7cls. Within the same computational budget of transformer blocks and a similar overall computational budget, POSTER-T outperforms image\_only case on both RAF-DB and AffectNet datasets. POSTER-T, POSTER-S, and POSTER have identical image backbone and landmark detector. The only difference between POSTER-T, POSTER-S, and POSTER is the number of transformer blocks. POSTER achieves the best results with 8 cross-fusion transformer blocks.   
\section{Transformer encoders depth:}
\label{depth}
\begin{figure*}[htp]
  \centering
  \includegraphics[width=1\linewidth]{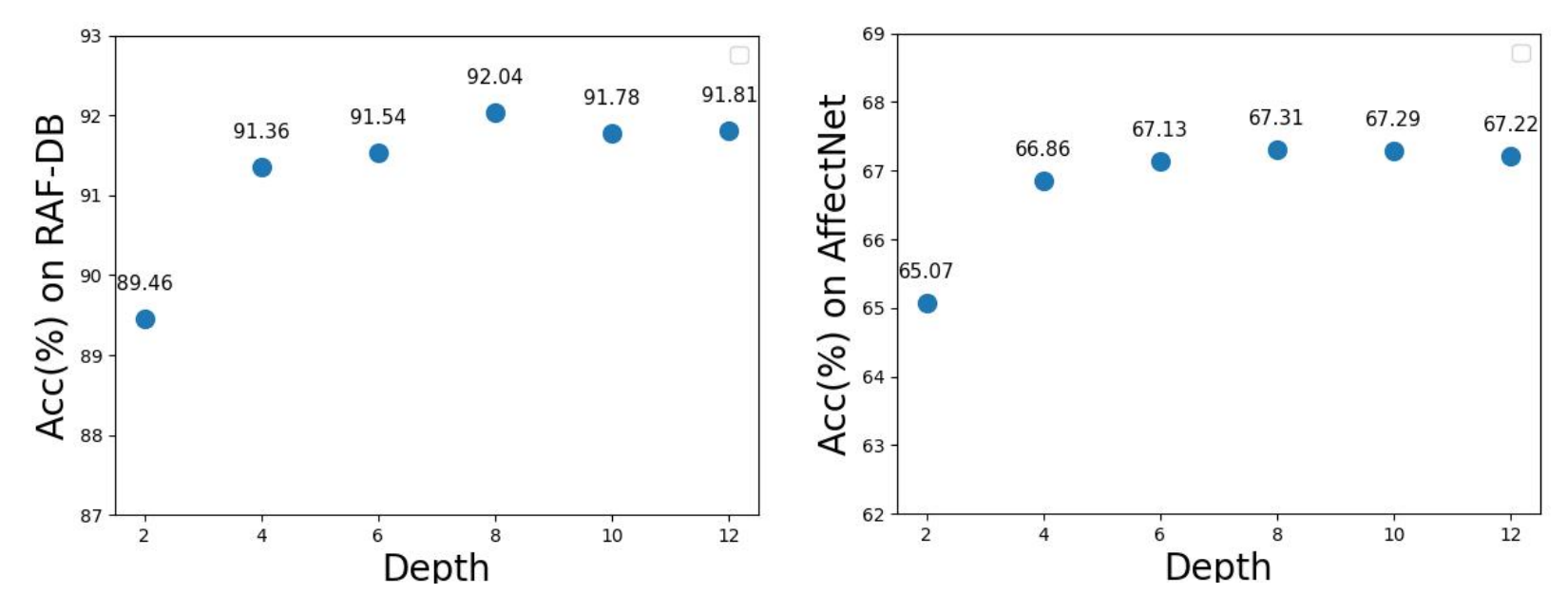}
  \vspace{-5pt}
  \caption{Evaluation of different numbers of transformer encoders (depth) on RAF-DB and AffectNet (7 cls) datasets.}
    \label{fig:ab_depth}
\end{figure*}

In Fig.~\ref{fig:ab_depth}, we plot relations between the accuracy with the network depth. When the number of transformer encoders is greater than 4, the performance is at a relatively high level on both RAF-DB and AffectNet datasets. We choose the depth number to be 8 in our final architecture since this provides the best results.

\begin{figure*}[htp]
  \centering
  \includegraphics[width=1\linewidth]{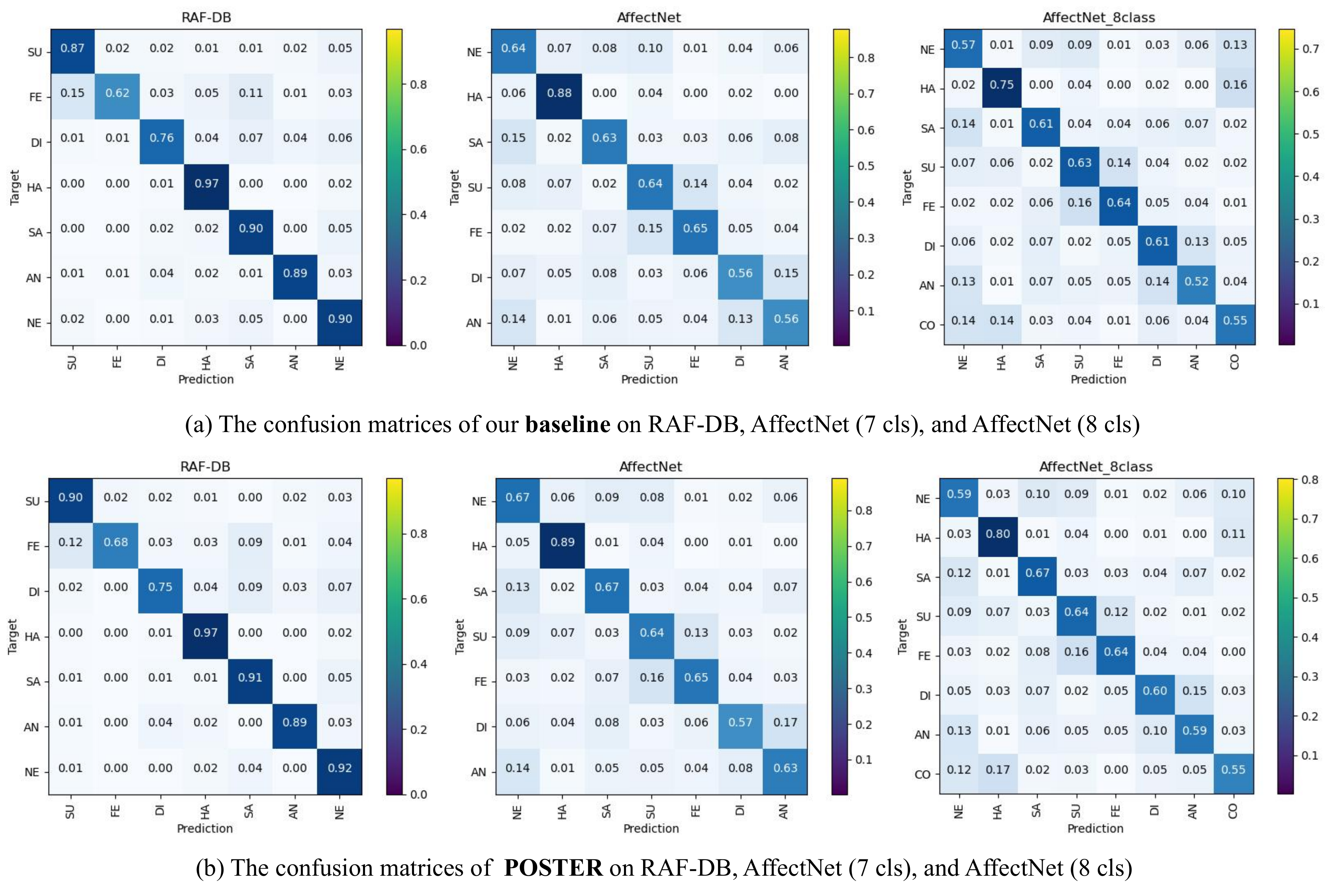}
  \vspace{-10pt}
  \caption{Confusion matrices of our baseline (a) and Poster (b) on RAF-DB \cite{rafdb}, AffectNet-7cls \cite{affectnet}, and AffectNet-8cls \cite{affectnet} datasets}
  \vspace{-5pt}
    \label{fig:matrix}
\end{figure*}

\section{Confusion Matrices:}
\label{confusion}
We show the confusion matrices of the baseline method and the proposed POSTER on RAF-DB, AffectNet (7 cls), and AffectNet (8 cls) datasets in Fig.~\ref{fig:matrix} (a) and (b). 

Compared with the baseline, POSTER significantly improves the class-wise accuracy (diagonals of each confusion matrix) on all three experiments in Fig.~\ref{fig:matrix} which indicates that POSTER reduces intra-class discrepancy for FER. 
Given the target categories, the error rate of predicting into wrong categories also decreases most of the cases when comparing the same positions (except diagonals of each confusion matrix) between Fig.~\ref{fig:matrix} (a) and (b), which shows that POSTER alleviates inter-class similarity for FER. 

\clearpage

{\small
\bibliographystyle{ieee_fullname}
\bibliography{egbib}
}

\end{document}